\documentclass[10pt,letterpaper,twocolumn]{article}
\usepackage[latin9]{inputenc}
\usepackage{mathrsfs}
\usepackage{amsmath}
\usepackage{amssymb}
\usepackage{graphicx}
\usepackage[unicode=true,pdfusetitle,
 bookmarks=false,
 breaklinks=true,pdfborder={0 0 1},backref=section,colorlinks=false]
 {hyperref}
\hypersetup{
 pagebackref=true,letterpaper=true,colorlinks}
\usepackage{breakurl}

\makeatletter

\providecommand{\tabularnewline}{\\}

\usepackage{wacv}\usepackage{times}\usepackage{epsfig}

\wacvfinalcopy % *** Uncomment this line for the final submission

\def\wacvPaperID{277} % *** Enter the CVPR Paper ID here

\ifwacvfinal\fi

\makeatother

\begin{document}
%%%%%%%%% TITLE

\title{Improved Deep Learning of Object Category using Pose Information}

\author{Jiaping Zhao, 
 Laurent Itti \\
 University of Southern California\\
 %Institution1 address\\
\texttt{\small{}{}\{jiapingz, itti\}@usc.edu}{\small{}{} }}
\maketitle
\begin{abstract}
Despite significant recent progress, the best available computer vision algorithms still lag far behind human
capabilities, even for recognizing individual discrete objects under various poses, illuminations, and backgrounds. Here
we present a new approach to using object pose information to improve deep network learning. While existing large-scale
datasets, e.g. ImageNet, do not have pose information, we leverage the newly published turntable dataset, iLab-20M,
which has $\sim$22M images of 704 object instances shot under different lightings, camera viewpoints and turntable
rotations, to do more controlled object recognition experiments. We introduce a new convolutional neural network
architecture, what/where CNN (2W-CNN), built on a linear-chain feedforward CNN (e.g., AlexNet), augmented by
hierarchical layers regularized by object poses. Pose information is only used as feedback signal during training, in
addition to category information, but is not needed during test. To validate the approach, we train both 2W-CNN and
AlexNet using a fraction of the dataset, and 2W-CNN achieves $6\%$ performance improvement in category prediction. We
show mathematically that 2W-CNN has inherent advantages over AlexNet under the stochastic gradient descent (SGD)
optimization procedure. Furthermore, we fine-tune object recognition on ImageNet by using the pretrained 2W-CNN and
AlexNet features on iLab-20M, results show significant improvement compared with training AlexNet from
scratch. Moreover, fine-tuning 2W-CNN features performs even better than fine-tuning the pretrained AlexNet
features. These results show that pretrained features on iLab-20M generalize well to natural image datasets, and 2W-CNN
learns better features for object recognition than AlexNet.
\end{abstract}
%%%%%%%%% BODY TEXT

\section{Introduction}

\begin{figure}
\centering{}\includegraphics[width=0.47\textwidth]{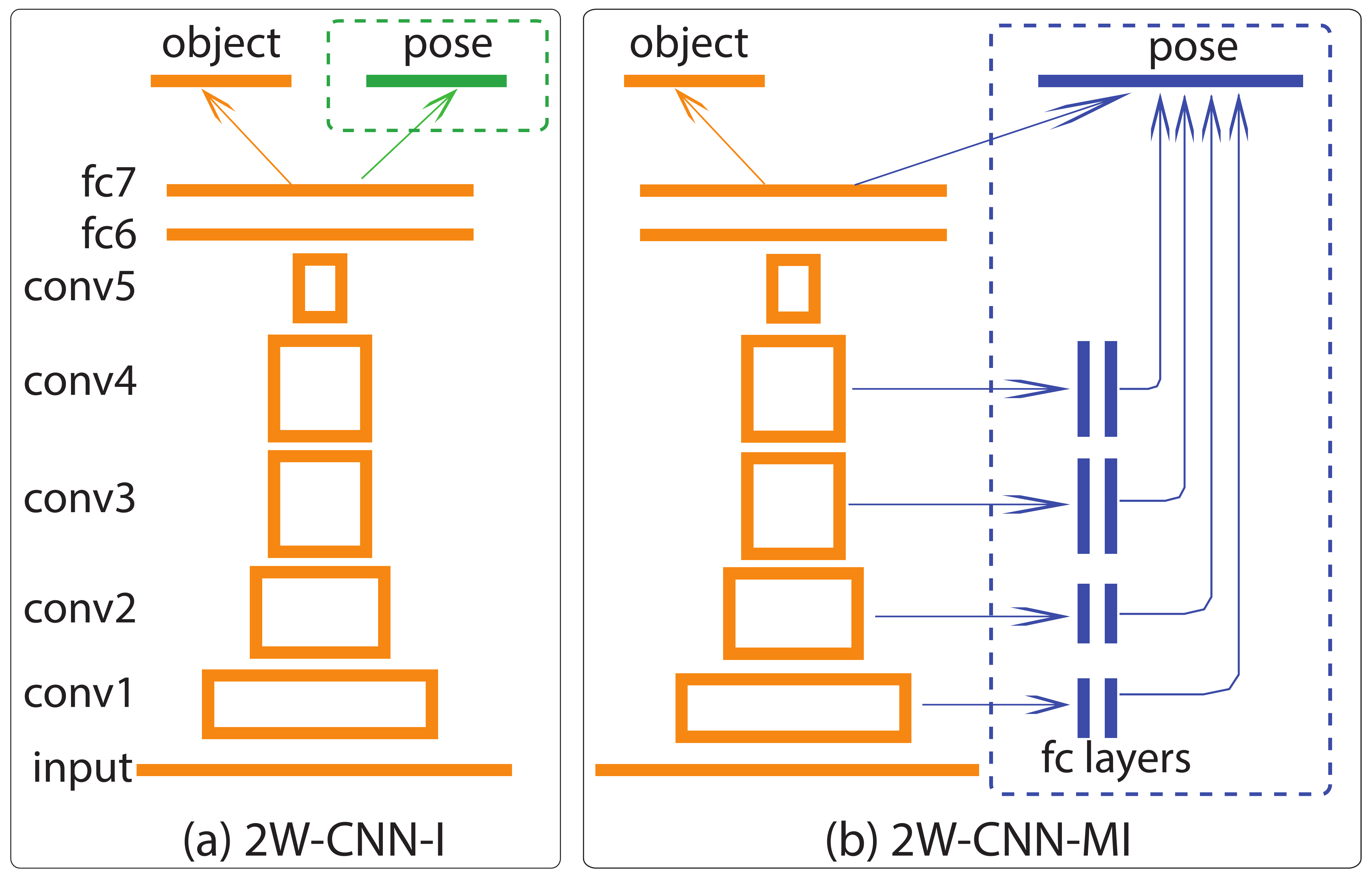} 

\caption{\label{fig:arc}2W-CNN architecture. The orange architecture is
 AlexNet, and we build two what/where convolutional neural
network architectures from it: (a) 2W-CNN-I: object pose information
(where) is linked to the top fully connected layer (fc7) only; (b)
2W-CNN-MI: object pose labels have direct pathways to all convolutional
layers. The additionally appended pose architectures (green in (a)
and blue in (b)) are used in training to regularize the deep feature
learning process, and in testing, we prune them and use the remaining
AlexNet for object recognition (what). Hence, although feedforward
connection arrows are shown, all blue and green connections are only
used for backpropagation.}
\end{figure}

%\vspace{-1em}

Deep convolutional neural networks (CNNs) have achieved great success
in image classification \cite{krizhevsky2012imagenet,szegedy2014going},
object detection \cite{sermanet2013overfeat,girshick2014rich}, image
segmentation \cite{chen2014semantic}, activity recognition \cite{karpathy2014large,simonyan2014two}
and many others. Typical CNN architectures, including AlexNet \cite{krizhevsky2012imagenet}
and VGG \cite{DBLP:journals/corr/SimonyanZ14a}, consist of several
stages of convolution, activation and pooling, in which pooling subsamples
feature maps, making representations locally translation invariant.
After several stages of pooling, the high-level feature representations
are invariant to object pose over some limited range, which is generally
a desirable property. Thus, these CNNs only preserve ``what'' information
but discard ``where'' or pose information through the multiple stages
of pooling. However, as argued by Hinton et al.~\cite{hinton2011transforming},
artificial neural networks could use local ``capsules'' to encapsulate
both ``what'' and ``where'' information, instead of using a single
scalar to summarize the activity of a neuron. Neural architectures
designed in this way have the potential to disentangle visual entities
from their instantiation parameters \cite{ranzato2007unsupervised,zhao2015stacked,goroshin2015learning}.

In this paper, we propose a new deep architecture built on a traditional
ConvNet (AlexNet, VGG), but with two label layers, one for category
(what) and one for pose (where; Fig.~$\ref{fig:arc}$). We name this
a what/where convolutional neural network (2W-CNN). Here, object category
is the class that an object belongs to, and pose denotes any factors
causing objects from the same class to have different appearances
on the images. This includes camera viewpoint, lighting, intra-class
object shape variances, etc. By explicitly adding pose labels to the
top of the network, 2W-CNN is forced to learn multi-level feature
representations from which both object categories and pose parameters
can be decoded. 2W-CNN only differs from traditional CNNs during training:
two streams of error are backpropagated into the convolutional layers,
one from category and the other from pose, and they jointly tune the
feature filters to simultaneously capture variability in both category
and pose. When training is complete, we prune all the auxiliary layers
in 2W-CNN, leaving only the base architecture (traditional ConvNet,
with same number of degrees of freedom as the original), and we use
it to predict the category label of a new input image. By explicitly
incorporating ``where'' information to regularize the feature learning
process, we experimentally show that the learned feature representations
are better delineated, resulting in better categorization accuracy.

\section{Related work}

This work is inspired by the concept revived by Hinton et al.~\cite{hinton2011transforming}.
They introduced ``capsules'' to encapsulate both ``what'' and
``where'' into a highly informative vector, and then feed both to
the next layer. In their work, they directly fed translation/transformation
information between input and output images as known variables into
the auto-encoders, and this essentially fixes ``where'' and forces
``what'' to adapt to the fixed ``where''. In contrast, in our
2W-CNN, ``where'' is an output variable, which is only used to back-propagate
errors. It is never fed forward into other layers as known variable.
In \cite{zhao2015stacked}, the authors proposed 'stacked what-where
auto-encoders' (SWWAE), which consists of a feed-forward ConvNet (encoder),
coupled with a feed-back DeConvnet (decoder). Each pooling layer from
the ConvNet generates two sets of variables, ``what'' which records
the features in the receptive field and is fed into the next layer,
and ``where'' which remembers the position of the interesting features
and is fed into the corresponding layer of the DeConvnet. Although
they explicitly build ``where'' variables into the architecture,
``where'' variables are always complementary to the ``what'' variables
and only help to record the max-pooling switch positions. In this
sense, ``where'' is not directly involved in the learning process.
In 2W-CNN, we do not have explicit ``what'' and ``where'' variables, instead, they are implicitly expressed
by neurons in the intermediate layers. Moreover, ``what'' and ``where''
variables from the top output layer are jointly engaged to tune filters
during learning. A recent work \cite{goroshin2015learning} proposes
a deep generative architecture to predict video frames. The representation
learned by this architecture has two components: a locally stable
``what'' component and a locally linear ``where'' component. Similar
to \cite{zhao2015stacked}, ``what'' and ``where'' variables are
explicitly defined as the output of 'max-pooling' and 'argmax-pooling'
operators, as opposed to our implicit 2W-CNN approach.

In \cite{wohlhart2015learning}, the authors propose to learn image
descriptors to simultaneously recognize objects and estimate their
poses. They train a linear chain feed-forward deep convolutional network
by including relative pose and object category similarity and dissimilarity
in their cost function, and then use the top layer output as image
descriptor. However, \cite{wohlhart2015learning} focus on learning
image descriptors, then recognizing category and pose through a nearest
neighbor search in descriptor space, while we investigate how explicit,
absolute pose information can improve category learning. \cite{bakry2014untangling}
introduces a method to separate manifolds from different categories
while being able to predict object pose. It uses HOG features as image
representations, which is known to be suboptimal compared to statistically
learned deep features, while we learn deep features with the aid of
pose information.

\begin{figure}[!htbp]
\begin{center}
\includegraphics [width=0.48\textwidth] {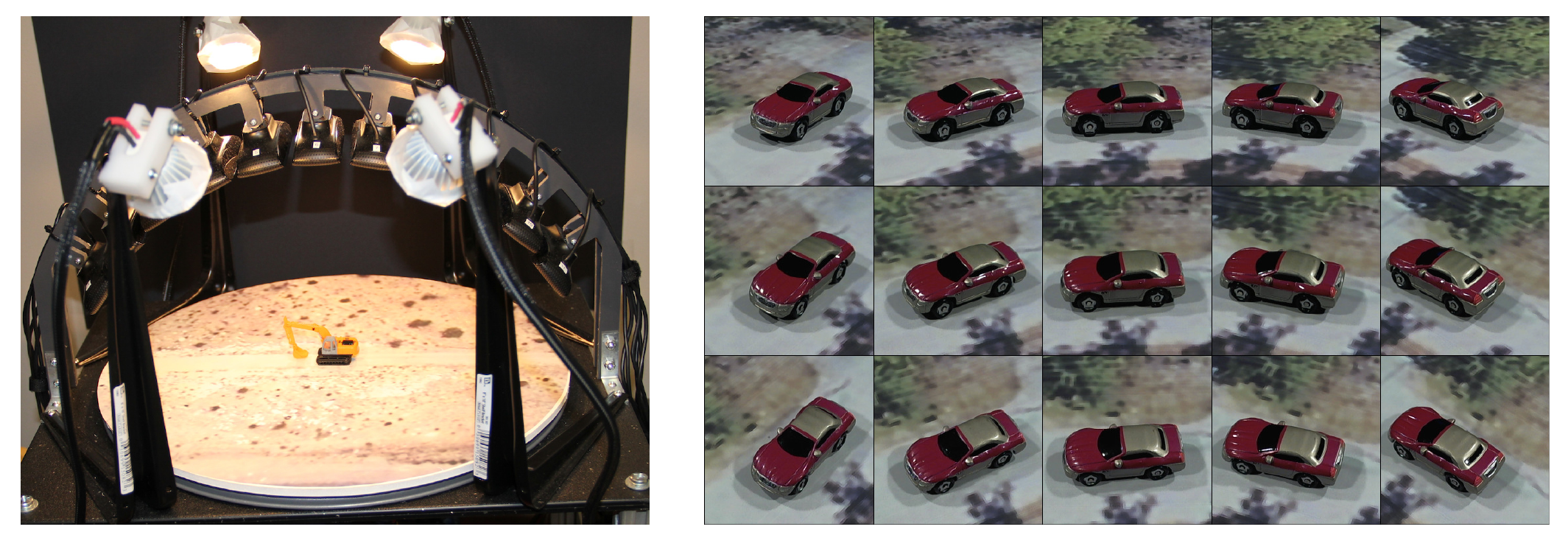}

\caption{\label{fig:Left:-turntable-setup}Left: turntable setup; Right: one
exemplar car shot under different viewpoints.}
\end{center}
\end{figure}

In sum, our architecture differs from the above in several aspects:
(1) 2W-CNN is a feed-forward discriminative architecture as opposed
to an auto-encoder; (2) we do not explicitly define ``what'' and
``where'' neurons, instead, they are
implicitly expressed by intermediate neurons; (3) we use explicit,
absolute pose information, only during back-propagation, and not in
the feed-forward pass.

Our architecture, 2W-CNN, also belongs to the framework of multi-task learning (MTL), where the basic notion is that
using a single network to learn two or more \textit{related} tasks yields better performance than using one dedicated
network for each task \cite{caruana1997multitask,baxter2000model,pan2010survey}.  Recently, several efforts have
explored multi-task learning using deep neural networks, for face detection, phoneme recognition, scene classification
and pose estimation \cite{zhang2014facial,zhang2014improving,seltzer2013multi,huang2013multi,su2015render}.  All of them
use a similar linear feed-forward architecture, with all task label layers appended onto the last fully connected
layer. In the end, all tasks in these applications share the same representations.  Although they
\cite{seltzer2013multi,zhang2014facial} do differentiate principal and auxiliary tasks by assigning larger/smaller
weights to principal/auxiliary task losses in the objective function, they never make a distinction of tasks when
designing the deep architecture.  Our architecture, 2W-CNN-I (see \ref{2W-CNN-I-arc} for definition), is similar to
theirs, however, 2W-CNN-MI (see \ref{2W-CNN-MI-arc} for definition) is very different: pose is the auxiliary task, and
it is designed to support the learning of the principal task (object recognition) at multiple levels. Concretely,
auxiliary labels (pose) have direct pathways to all convolutional layers, such that features in the intermediate layers
can be directly regularized by the auxiliary task. We experimentally show that 2W-CNN-MI, which embodies a new kind of
multi-task learning, is superior to 2W-CNN-I for object recognition, and this indicates that 2W-CNN-MI is advantageous
to the previously published deep multi-task learning architectures.

\section{A brief introduction of the iLab-20M dataset}

\noindent iLab-20M \cite{Borji_2016_CVPR} was collected by hypothesizing that training can
be greatly improved by using many different views of different instances
of objects in a number of categories, shot in many different environments,
and with pose information explicitly known. Indeed, biological systems
can rely on object persistence and active vision to obtain many different
views of a new physical object. In monkeys, this is believed to be
exploited by the neural representation \cite{li2010unsupervised},
though the exact mechanism remains poorly understood. 

iLab-20M is a turntable dataset,
with settings as follows: the turntable consists of a 14''-diameter circular plate
actuated by a robotic servo mechanism. A CNC-machined semi-circular
arch (radius 8.5'') holds 11 Logitech C910 USB webcams which
capture color images of the objects placed on the turntable. A micro-controller
system actuates the rotation servo mechanism and switches on and off
4 LED lightbulbs. Lights are
controlled independently, in 5 conditions: all lights on, or one of
the 4 lights on.

Objects were mainly Micro Machines toys (Galoob Corp.) and
N-scale model train toys. These objects present the advantage of small scale, yet demonstrate
a high level of detail and, most remarkably, a wide range of shapes
(i.e., many different molds were used to create the objects, as opposed
to just a few molds and many different painting schemes). Backgrounds
were 125 color printouts of satellite imagery from the Internet. Every object was shot on at least 14 backgrounds, in
a relevant context (e.g., cars on roads, trains on railtracks, boats
on water).

In total, 1,320 images were captured for each object and background
combination: 11 azimuth angles (from the 11 cameras), 8 turntable
rotation angles, 5 lighting conditions, and 3 focus values (-3, 0,
and +3 from the default focus value of each camera). Each image was
saved with lossless PNG compression ($\sim$1 MB per image). The complete
dataset hence consists of 704 object instances (15 categories), each shot on 14 or more backgrounds,
with 1,320 images per object/background combination, or almost 22M
images. The dataset is freely available and distributed on several
hard drives. One exemplar car instance shot under different viewpoints
are shown in Fig. \ref{fig:Left:-turntable-setup}(right).

\section{Network Architecture and its Optimization}

\noindent In this section, we introduce our new architecture, 2W-CNN,
and some properties of its critical points achieved under the stochastic
gradient descent (SGD) optimization.

\subsection{Architecture}

Our architecture, 2W-CNN, can be built on any of the CNNs, and, here, without loss of generality, we use AlexNet
\cite{krizhevsky2012imagenet} (but see Supplementary Materials for results using VGG as well).
iLab-20M has detailed pose information for each image. In the testing presented here, we only consider 10 categories,
and the 8 turntable rotations and 11 camera azimuth angles, i.e., 88 discrete poses. It would be straightforward to use
more categories and take light source, focus, etc.~into account as well.

Our building base, AlexNet, is adapted here to be suitable for our dataset: two changes are made, compared to AlexNet in \cite{krizhevsky2012imagenet} (1) we change the number of units on fc6 and fc7 from 4096 to 1024, since we only have ten categories here; (2) we append a batch normalization layer after each convolution layer (see the supplementary materials for architecture specifications).

%[[[put in suppl]]]
%We use AlexNet re-implemented in \cite{vedaldi15matconvnet}, whose
%architecture is: the first two layers are divided into 4 sub-layers:
%convolution, local response normalization (LRN), ReLUs and max-pooling.
%Layer 3 and 4 are composed of convolution and ReLUs. Layer 5 consists
%of convolution, followed by ReLUs and max-pooling. There are two fully
%connected layers, fc6 and fc7, stacked on top of pool5. The last object
%label layer is appended onto fc7. The architecture specifications
%are listed in Table \ref{tab:architectures-of-2wcnn}. In our case,
%we change the number of units on fc6 and fc7 to be 1024, which is
%the only difference with widely used AlexNet on ImageNet and places
%databases (4096 units). The output object label layer has 10 units
%(10 object categories).
%[[[including the table]]]

We design two variants of our approach: (1) 2W-CNN-I (with I for {\em injection}), a what/where CNN with both pose and
category information injected into the top fully connected layer; (2) 2W-CNN-MI {\em(multi-layer injection)}, a
what/where CNN with category still injected at the top, but pose injected into the top and also directly into all 5
convolutional layers. Our motivation for multiple injection is as follows: it is generally believed that in CNNs, low-
and mid-level features are learned mostly in lower layers, while, with increasing depth, more abstract high-level
features are learned \cite{zeiler2014visualizing,zhou2014object}.  Thus, we reasoned that detailed pose information
might also be used differently by different layers. ``Multi-layer injection'' in 2W-CNN-MI is similar to skip connections in neural
networks. Skip connection is a more generic terminology, while 2W-CNN-MI uses a specific pattern of
skip connections designed specifically to make pose errors directly back propagate into lower layers. Our architecture
details are as follows.

\noindent \textbf{2W-CNN-I\label{2W-CNN-I-arc}} is built on AlexNet,
and we further append a pose layer (88 neurons) to fc7. The architecture
is shown in Fig.~\ref{fig:arc}. 2W-CNN-I is trained to predict both
what and where. We treat both prediction tasks as classification problems,
and use softmax regression to define the individual loss. The total
loss is the weighted sum of individual losses:

%\vspace{-0.8em}

\begin{equation}
\mathcal{L}=\mathcal{L}(object)+\lambda\mathcal{L}(pose)\label{eq:loss-2w-cnn}
\end{equation}
where $\lambda$ is a balancing factor, set to 1 in experiments. Although
we do not have explicit what and where neurons in 2W-CNN-I, feature
representations (neuron responses) at fc7 are trained such that both
object category (what) and pose information (where) can be decoded;
therefore, neurons in fc7 can be seen to have implicitly encoded what/where
information. Similarly, neurons in intermediate layers also implicitly
encapsulate what and where, since both pose and category errors at
the top are back-propagated consecutively into all layers, and features
learned in the low layers are adapted to both what and where.

\noindent \textbf{2W-CNN-MI} \label{2W-CNN-MI-arc}is built on AlexNet
as well, but in this variant we add direct pathways from each convolutional
layer to the pose layer, such that feature learning at each convolutional
layer is directly affected by pose errors (Fig.~\ref{fig:arc}).
Concretely, we append two fully connected layers to each convolutional
layer, including pool1, pool2, conv3 and conv4, and then add a path
from the 2nd fully connected layer to the pose category layer. Fully
connected layers appended to pool1 and pool2 have 512 neurons and
those appended to conv3 and conv4 have 1024 neurons. At last, we directly
add a pathway from fc7 to the pose label layer. The reason we do not
append two additional fully connected layers to pool5 is that the
original AlexNet already has fc6 and fc7 on top of pool5; thus, our
fc6 and fc7 are shared by the object category layer and the pose layer.
%The architecture specifications are listed in Table \ref{tab:architectures-of-2wcnn}.

The loss function of 2W-CNN-MI is the same as that of of 2W-CNN-I
(Eq.~\ref{eq:loss-2w-cnn}). In 2W-CNN-MI, activations from 5 layers,
namely pool1fc2, pool2fc2, conv3fc2, conv4fc2 and fc7, are all fed
into the pose label layer, and responses at the pose layer are the
accumulated activations from all 5 pathways, i.e.,

\vspace{-1em}

\begin{equation}
a(\mathscr{L}_{P})=\sum_{l}a(l)\cdot\mathcal{W}_{l-\mathscr{L}_{P}}\label{eq:pose-activations}
\end{equation}
where $l$ is one from those 5 layers, $\mathcal{W}_{l-\mathscr{L}_{P}}$
is the weight matrix between $l$ and pose label layer $\mathscr{L}_{P}$,
and $a(l)$ are feature activations at layer $l$.

\subsection{Optimization\label{sub:Optimization}}

We use stochastic gradient descent (SGD) to minimize the loss function. In practice, it either finds
a local optimum or a saddle point \cite{dauphin2014identifying,pascanu2014saddle}
for non-convex optimization problems. How to escape a saddle point
or reach a better local optimum is beyond the scope of our work here.
Since both 2W-CNN and AlexNet are optimized using SGD, readers may
worry that object recognition performance differences between 2W-CNN
and AlexNet might be just occasional and depending on initializations,
while here we show theoretically that it is easier to find a better
critical point in the parameter space of 2W-CNN than in the parameter
space of AlexNet by using SGD.

\begin{figure}
\begin{centering}
\includegraphics[width=0.18\textwidth]{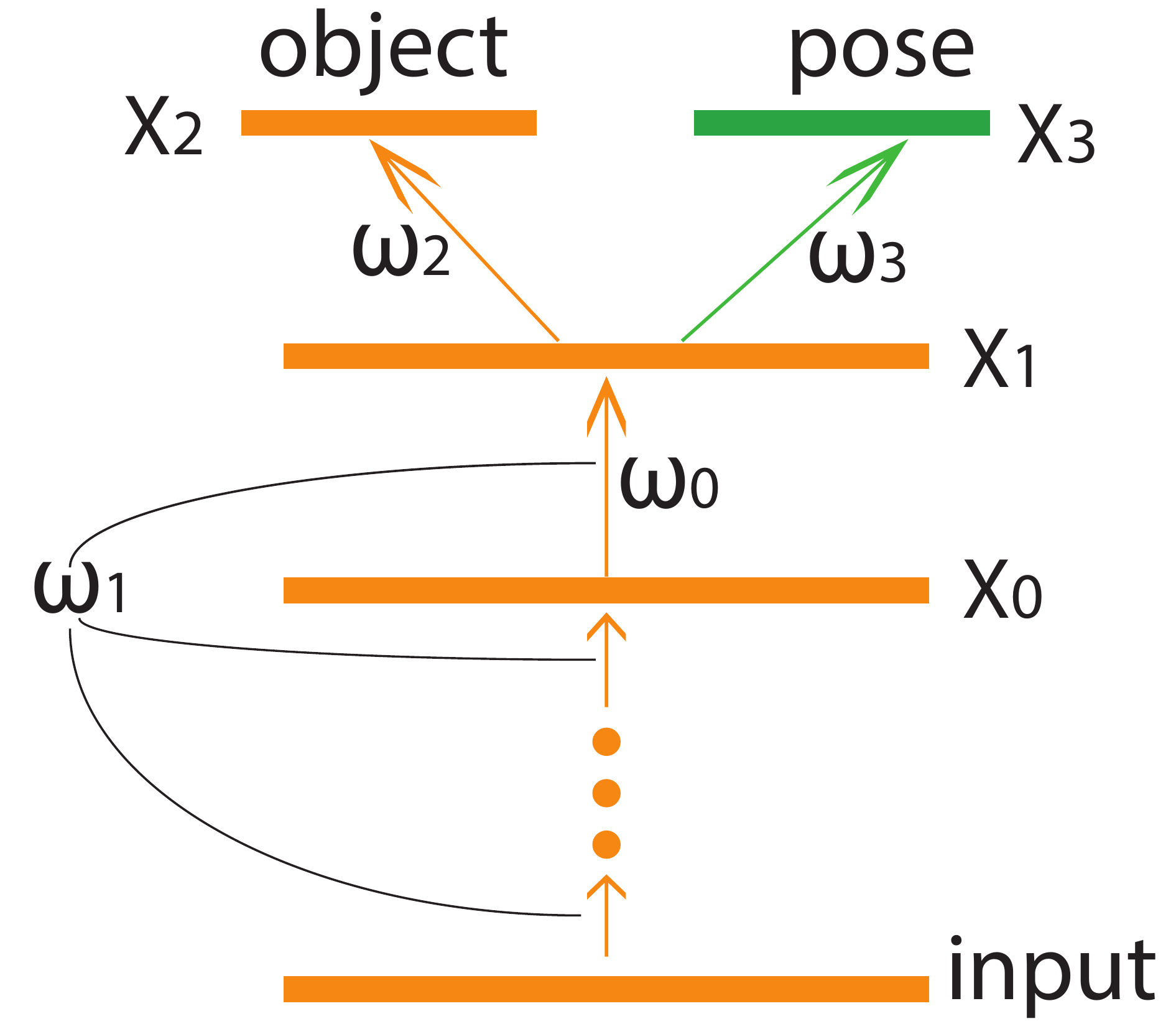} 
\par\end{centering}

\centering{}\caption{\label{fig:proof}A simplified CNN used in proof.}
\end{figure}

We prove that, in practice, a critical point of AlexNet is not a critical
point of 2W-CNN, while a critical point of 2W-CNN is a critical point
of AlexNet as well. Thus, if we initialize weights in a 2W-CNN from
a trained AlexNet (i.e., we initialize $\omega_{1},\omega_{2}$ from
the trained AlexNet, while initializing $\omega_{3}$ by random Gaussian
matrices in Fig.~\ref{fig:proof}), and continue training 2W-CNN
by SGD, parameter solutions will gradually step away from the initial
point and reach a new (better) critical point. However, if we initialize
parameters in AlexNet from a trained 2W-CNN and continue training,
the parameter gradients in AlexNet at the initial point are already
near zero and no better critical point is found. Indeed, in the next
section we verify this experimentally.

Let $\mathcal{L}=f(\omega_{1},\omega_{2})$ and $\mathcal{\hat{L}}=f(\omega_{1},\omega_{2})+g(\omega_{1},\omega_{3})$
be the softmax loss functions of AlexNet and 2W-CNN respectively,
where $f(\omega_{1},\omega_{2})$ in $\mathcal{L}$ and $\mathcal{\hat{L}}$
are exactly the same, we show in practical cases that: (1) if $(\omega_{1}^{\prime},\omega_{2}^{\prime})$
is a critical point of $\mathcal{L}$, then $(\omega_{1}^{\prime},\omega_{2}^{\prime},\omega_{3})$
is not a critical point of $\mathcal{\hat{L}}$; (2) on the contrary,
if $(\omega_{1}^{\prime\prime},\omega_{2}^{\prime\prime},\omega_{3}^{\prime\prime})$
is a critical point of $\hat{\mathcal{L}}$, $(\omega_{1}^{\prime\prime},\omega_{2}^{\prime\prime})$
is a critical point of $\mathcal{L}$ as well. Here we prove (1) but
refer the readers to the supplementary materials for the proof of
(2).

\vspace{-1.2em}

\begin{equation}
\left.\begin{array}{c}
\frac{\partial\mathcal{L}}{\partial\omega_{1}^{\prime}}=\frac{\partial f}{\partial\omega_{1}^{\prime}}=\overrightarrow{0}\\
\frac{\partial\mathcal{L}}{\partial\omega_{2}^{\prime}}=\frac{\partial f}{\partial\omega_{2}^{\prime}}=\overrightarrow{0}
\end{array}\right\} \rightsquigarrow\left\{ \begin{array}{c}
\frac{\partial\mathcal{\hat{L}}}{\partial\omega_{1}^{\prime}}=\frac{\partial f}{\partial\omega_{1}^{\prime}}+\frac{\partial g}{\partial\omega_{1}^{\prime}}\neq\overrightarrow{0}\\
\frac{\partial\mathcal{\hat{L}}}{\partial\omega_{2}^{\prime}}=\frac{\partial f}{\partial\omega_{2}^{\prime}}=\overrightarrow{0}\\
\frac{\partial\mathcal{\hat{L}}}{\partial\omega_{3}}=\frac{\partial g}{\partial\omega_{3}}\neq\overrightarrow{0}
\end{array}\right.\label{eq:alexnet22w-cnn}
\end{equation}

%\vspace{-2em}

\noindent \textbf{Proof:} assume there is at least one non-zero entry
in $x_{1}\in\mbox{\ensuremath{\mathcal{R}}}^{1024}$ (in practice
$x_{1}\neq\overrightarrow{0}$, $x_{0}\neq\overrightarrow{0}$, at
least one non-zero entry, see supplementary materials), and let $x_{1}^{nz}$
be that non-zero element ($nz$ is its index, $nz\in\{1,2,...,1024\}$).
Without loss of generality, we initialize all entries in $\omega_{3}\in\mathcal{R}^{88\times1024}$
be to 0, except one entry $\omega_{3}^{nz,1}$, which is the weight
between $x_{1}^{nz}$ and $x_{3}^{1}$. Since we have $x_{3}=\omega_{3}x_{1},\:(x_{3}=\mathcal{R}^{88})$,
then $x_{3}^{1}=\omega_{3}^{nz,1}\cdot x_{1}^{nz}\neq0$, while $x_{3}^{i}=0\:(i\neq1)$.

In the case of softmax regression, we have: $\frac{\partial\mathcal{\hat{L}}}{\partial x_{3}^{c}}=1-e^{x_{3}^{c}}/\sum_{i}e^{x_{3}^{i}}$;
$\frac{\partial\mathcal{\hat{L}}}{\partial x_{3}^{i}}=-e^{x_{3}^{i}}/\sum_{i}e^{x_{3}^{i}}$
when $i\neq c$, where $x_{3}^{i}$ is the $i^{th}$ entry in the
vector $x_{3}$, and $c$ is the index of the ground truth pose label.
Since $x_{3}^{1}\neq0$ and $x_{3}^{i}=0(i\neq1)$, $\frac{\partial\mathcal{\hat{L}}}{\partial x_{3}^{i}}\neq0\:(i\in\{1,2,...,88\})$.
By chain rule, we have $\frac{\partial\mathcal{\hat{L}}}{\partial\omega_{3}^{mn}}=\frac{\partial\mathcal{\hat{L}}}{\partial x_{3}^{n}}\cdot\frac{\partial x_{3}^{n}}{\partial\omega_{3}^{mn}}=\frac{\partial\mathcal{\hat{L}}}{\partial x_{3}^{n}}\cdot x_{1}^{m}$,
and therefore, as long as not all entries in $x_{1}$ are 0, i.e.,
$x_{1}\neq\overrightarrow{0}$, we have $\frac{\partial\mathcal{\hat{L}}}{\partial\omega_{3}}=\frac{\partial\mathcal{\hat{L}}}{\partial x_{3}}\cdot x_{1}\neq\overrightarrow{0}$.

To show $\frac{\partial\mathcal{\hat{L}}}{\partial\omega_{1}^{\prime}}=\frac{\partial f}{\partial\omega_{1}^{\prime}}+\frac{\partial g}{\partial\omega_{1}^{\prime}}\neq\overrightarrow{0}$,
we only have to show $\frac{\partial g}{\partial\omega_{1}^{\prime}}\neq\overrightarrow{0}$
(since $\frac{\partial f}{\partial\omega_{1}^{\prime}}=\overrightarrow{0}$).
By chain rule, we have $\frac{\partial g}{\partial\omega_{1}^{\prime}}=\frac{\partial g}{\partial x_{1}}\cdot\frac{\partial x_{1}}{\partial\omega_{1}^{\prime}}$,
where $\frac{\partial g}{\partial x_{1}^{i}}=0$, when $i\neq nz$;
$\frac{\partial g}{\partial x_{1}^{nz}}=\omega_{3}^{nz,1}-\omega_{3}^{nz,1}\cdot e^{x_{3}^{1}}/\sum_{i}e^{x_{3}^{i}}\neq0$.
Let the weight matrix between $x_{0}$ and $x_{1}$ be $\omega_{0}^{\prime}$,
and by definition, $\omega_{0}^{\prime}\in\omega_{1}^{\prime}$. Now
we have $\frac{\partial x_{1}^{nz}}{\partial\omega_{0}^{\prime}}\neq\overrightarrow{0}$,
otherwise $x_{0}=\overrightarrow{0}$. Therefore $\frac{\partial g}{\partial\omega_{1}^{\prime}}=\frac{\partial g}{\partial x_{1}}\cdot\frac{\partial x_{1}}{\partial\omega_{1}^{\prime}}\neq\overrightarrow{0}$,
since $\frac{\partial g}{\partial x_{1}}\neq\overrightarrow{0}$ and
$\frac{\partial x_{1}}{\partial\omega_{1}^{\prime}}\neq\overrightarrow{0}$.

However, a critical point of 2W-CNN is a critical point of AlexNet,
i.e., Eq.~\ref{eq:2w-cnn2alexnet}, see supplementary materials for
the proof, and next section for experimental validation.

\vspace{-1.2em}

\begin{equation}
\left.\begin{array}{c}
\frac{\partial\mathcal{\hat{L}}}{\partial\omega_{1}^{\prime\prime}}=\frac{\partial f}{\partial\omega_{1}^{\prime\prime}}+\frac{\partial g}{\partial\omega_{1}^{\prime\prime}}=\overrightarrow{0}\\
\frac{\partial\mathcal{\hat{L}}}{\partial\omega_{2}^{\prime\prime}}=\frac{\partial f}{\partial\omega_{2}^{\prime\prime}}=\overrightarrow{0}\\
\frac{\partial\mathcal{\hat{L}}}{\partial\omega_{3}^{\prime\prime}}=\frac{\partial g}{\partial\omega_{3}^{\prime\prime}}=\overrightarrow{0}
\end{array}\right\} \rightsquigarrow\left\{ \begin{array}{c}
\frac{\partial\mathcal{L}}{\partial\omega_{1}^{\prime\prime}}=\frac{\partial f}{\partial\omega_{1}^{\prime\prime}}=\overrightarrow{0}\\
\frac{\partial\mathcal{L}}{\partial\omega_{2}^{\prime\prime}}=\frac{\partial f}{\partial\omega_{2}^{\prime\prime}}=\overrightarrow{0}
\end{array}\right.\label{eq:2w-cnn2alexnet}
\end{equation}

\section{Experiments}

%\vspace{-0.5em}

In experiments, we demonstrate the effectiveness of 2W-CNN for object
recognition against linear-chain deep architectures (e.g., AlexNet)
using the iLab-20M dataset. We do both quantitative comparisons and
qualitative evaluations. Further more, to show the learned features
on iLab-20M are useful for generic object recognitions, we adop the
``pretrain - fine-tuning'' paradigm, and fine tune object recognition
on the ImageNet dataset \cite{deng2009imagenet} using the pretrained
AlexNet and 2W-CNN-MI features on the iLab-20M dataset.

\subsection{Dataset setup}

\noindent \textbf{Object categories:} we use 10 (out of 15) categories
of objects in our experiments (Fig.~\ref{fig:t-SNE-visualization}),
and, within each category, we randomly use 3/4 instances as training
data, and the remaining 1/4 instances for testing. Under this partition,
instances in test are never seen during training, which minimizes
the overlap between training and testing.

\noindent \textbf{Pose:} here we take images shot under one fixed
light source (with all 4 lights on) and camera focus (focus = 1),
but all 11 camera azimuths and all 8 turntable rotations (88 poses).

We end up with 0.65M (654,929) images in the training set and 0.22M
(217,877) in the test set. Each image is associated with 1 (out of
10) object category label and 1 (out of 88) pose label.

\subsection{CNNs setup}

We train 3 CNNs, AlexNet, 2W-CNN-I and 2W-CNN-MI, and compare their
performances on object recognition. We use the same initialization
for their common parameters: we first initialize AlexNet with random
Gaussian weights, and re-use these weights to initialize the AlexNet
component in 2W-CNN-I and 2W-CNN-MI. We then randomly initialize the
additional parameters in 2W-CNN-I / 2W-CNN-MI.

No data augmentation: to train AlexNet for object recognition, in
practice, one often takes random crops and also horizontally flips
each image to augment the training set. However, to train 2W-CNN-I
and 2W-CNN-MI, we could take random crops but we should not horizontally
flip images, since flipping creates a new unknown pose. For a fair
comparison, we do not augment the training set, such that all 3 CNNs
use the same amounts of images for training.

Optimization settings: we run SGD to minimize the loss function, but
use different starting learning and dropout rates for different CNNs.
AlexNet and 2W-CNN-I have similar amounts of parameters, while 2W-CNN-MI
has 15 times more parameters during training (but remember that all
three models have the exact same number of parameters during test).
To control overfitting, we use a smaller starting learning rate (0.001)
and a higher dropout rate (0.7) for 2W-CNN-MI, while for AlexNet and
2W-CNN-I, we set the starting learning rate and dropout rate to be
0.01 and 0.5. Each network is trained for 30 epochs, and approximately
150,000 iterations. To further avoid any training setup differences,
within each training epoch, we fix the image order. We train CNNs
using the publicly available Matconvnet \cite{vedaldi15matconvnet}
toolkit on a Nvidia Tesla K40 GPU.

\subsection{Performance evaluation}

In this section, we evaluate object recognition performance of the
3 CNNs. As mentioned, for both 2W-CNN-I and 2W-CNN-MI, object pose
information is only used in training, but the associated machinery
is pruned away before test (Fig.~\ref{fig:arc}).

Since all three architectures are trained by SGD, the solutions depend
on initializations. To alleviate the randomness of SGD, we run SGD
under different initializations and report the mean accuracy. We repeat
the training of 3 CNNs under 5 different initializations, and report
their mean accuracies and standard deviations in Table \ref{tab:Object-recognition-performances}.
Our main result is: (1) 2W-CNN-MI and 2W-CNN-I outperform AlexNet
by $6\%$ and $5\%$; (2) 2W-CNN-MI further improves the accuracy
by $1\%$ compared with 2W-CNN-I. This shows that, under the regularization
of additional pose information, 2W-CNN learns better deep features
for object recognition.

%\vspace{-1.5em}

\begin{table} [htbp]
\begin{center}
\newcommand{\tabincell}[2]{\begin{tabular}{@{}#1@{}}#2\end{tabular}}
\begin{tabular}{|c|c|c|c|}
\hline 
 & AlexNet & 2W-CNN-I & 2W-CNN-MI\tabularnewline
\hline 
\hline 
accuracy & \tabincell{c} {0.785\\($\pm$0.0019)} & \tabincell{c} {0.837 \\($\pm$0.0022)} & \tabincell{c} {$\textbf{0.848}$ \\ ($\pm$0.0031)} \tabularnewline
\hline 
mAP  & 0.787  & 0.833  & $\textbf{0.850}$\tabularnewline
\hline 
\end{tabular}
\caption{\label{tab:Object-recognition-performances}Object recognition performances
of AlexNet, 2W-CNN-I and 2W-CNN-MI on iLab-20M dataset. 2W-CNN-MI performed significantly better than AlexNet (t-test, $p<1.6\cdot10^{-5}$), and 2W-CNN-I ($p<6.3\cdot10^{-5}$) as well. 2W-CNN-MI was also significantly better than 2W-CNN-I ($p<.013$).}
\end{center}
\end{table}

%\vspace{-2em}

\begin{figure}[htbp!]
\begin{center}
\includegraphics[width=0.48\textwidth]{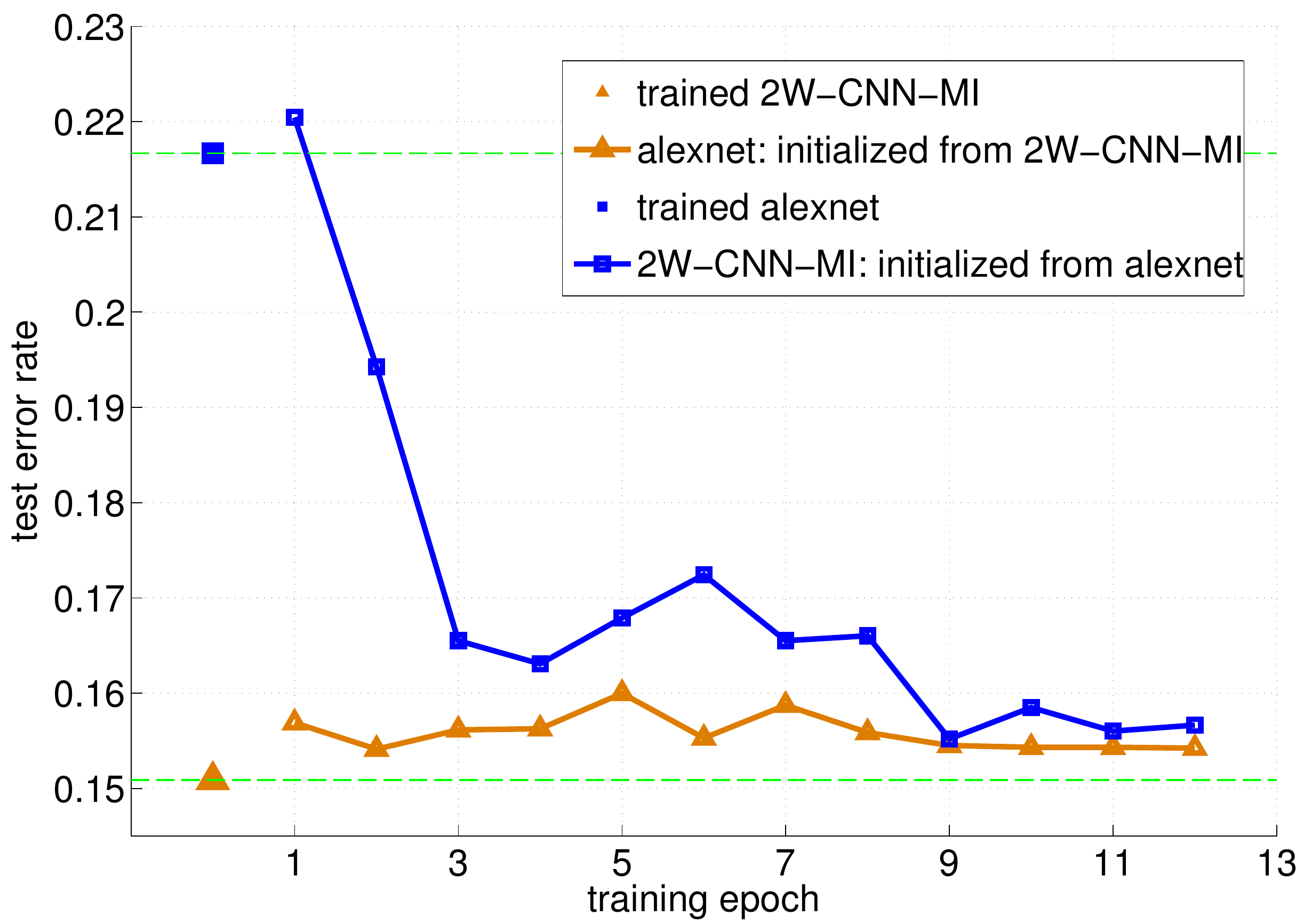} 
\par\end{center}%

\caption{\label{fig:warmstart}Critical points of CNNs. We initialize one network
from the trained other network, continue training, and record test
errors after each epoch. Starting from a critical point of AlexNet,
2W-CNN-MI steps away from it and reaches a new and better critical
point, while AlexNet initialized from 2W-CNN-MI fails to further improve
on test performance.}

\end{figure}

As proven in Sec.~\ref{sub:Optimization}, in practice a critical
point of AlexNet is not a critical point of 2W-CNN, while a critical
point of 2W-CNN is a critical point of AlexNet. We verify critical
points of two networks using experiments: (1) we use the trained AlexNet
parameters to initialize 2W-CNN, and run SGD to continue training
2W-CNN; (2) conversely, we initialize AlexNet from the trained 2W-CNN
and continue training for some epochs. We plot the object recognition
error rates on test data against training epochs in Fig.~\ref{fig:warmstart}.
As shown, 2W-CNN obviously reaches a new and better critical point
distinct from the initialization, while AlexNet stays around the same
error rate as the initialization.
%
%\begin{figure}
%\includegraphics[width=0.47\textwidth]{fig-warmstart}
%
%\caption{\label{fig:warmstart}Critical points of CNNs. We initialize one network
%from the trained other network, continue training, and record test
%errors after each epoch. Starting from a critical point of AlexNet,
%2W-CNN-MI steps away from it and reaches a new and better critical
%point, while AlexNet initialized from 2W-CNN-MI fails to further improve
%on test performance.}
%\end{figure}

%\begin{figure*}[htbp!]
%\begin{minipage}[c]{0.7\textwidth}%
%\begin{center}
%\includegraphics[width=0.9\textwidth]{fig-comple} 
%\par\end{center}%
%\end{minipage}%\hspace{0.1cm}
%\begin{minipage}[c]{0.30\linewidth}%
%\parbox[t]{1\textwidth}{%
%\caption{\label{fig:Complementarity}Complementarity of HOG-1D and DTW-MDS
%descriptors. HOG-1D is invariant to y-axis shift and insensitive to
%noise, while DTW-MDS is more contraction, stretching and translation
%invariant. The fusion of two descriptors benefits from their individual
%advantages, and outperforms each separate descriptor on most of the
%UCR datasets. Two plots in the 1st row show t-SNE \cite{van2008visualizing}
%visualization of HOG-1D and DTW-MDS descriptors of 246 subsequences.
%As we see from both plots, similar shaped subsequences are displayed
%proximately, indicating both descriptors capture shapes very well
%(Here, different subsequence colors are independent of t-SNE, but
%just for visual comfortability. 246 subsequences are k-means clustered
%into 10 clusters, with a random color assigned for each cluster).
%Two plots in the 2nd row show performance comparisons between HOG-1D+DTW-MDS
%and HOG-1D (DTW-MDS). On most datasets, the fused descriptor outperforms
%each separate one. By running Wilcoxon signed rank test, p-value between
%the fused and HOG-1D (DTW-MDS) is $6.7\cdot10^{-7}$ and $1.5\cdot10^{-5}$,
%showing fused descriptor performs significantly better.}
%%
%} %
%\end{minipage}
%\end{figure*}

\subsection{Decoupling of what and where}

2W-CNNs are trained to predict what and where. Although 2W-CNNs do
not have explicit what and where neurons, we experimentally show that
what and where information is implicitly expressed by neurons in different
layers.

One might speculate that in 2W-CNN, more than in standard AlexNet,
different units in the same layer might become either pose-defining
or identity-defining. A pose-defining unit should be sensitive to
pose, but invariant to object identity, and conversely. To quantify
this, we use entropy to measure the uncertainty of each unit to pose
and identity. We estimate pose and identity entropies of each unit
as follows: we use all test images as inputs, and we calculate the
activation ($a_{i}$) of each image for that unit. Then we compute
histogram distributions of activations against object category (10
categories in our case) and pose (88 poses), and let two distributions
be $\mathcal{P}_{obj}$ and $\mathcal{P}_{pos}$ respectively. The
entropies of these two distributions, $\mathscr{E}(\mathcal{P}_{obj})$
and $\mathscr{E}(\mathcal{P}_{pos})$, are defined to be the object
and pose uncertainty of that unit. For units to be pose-defining or
identity-defining, one entropy should be low and the other high, while
for units with identity and pose coupled together, both entropies
are high.

Assume there are $n$ units on some layer $l$ (e.g., 256 units on
pool5), each with identity and pose entropy $\mathscr{E}^{i}(obj)$
and $\mathscr{E}^{i}(pos)$ ($i\in\{1,2,...,n\}$), and we organize
$n$ identity entropies into a vector $\mathscr{E}(obj)=[\mbox{\ensuremath{\mathscr{E}}}^{1}(obj),\mbox{\ensuremath{\mathscr{E}}}^{2}(obj),...,\mbox{\ensuremath{\mathscr{E}}}^{n}(obj)]$
and $n$ pose entropies into a vector $\mathscr{E}(pos)=[\mbox{\ensuremath{\mathscr{E}}}^{1}(pos),\mbox{\ensuremath{\mathscr{E}}}^{2}(pos),...,\mbox{\ensuremath{\mathscr{E}}}^{n}(pos)]$.
If $n$ units are pose/identity decoupled, then $\mathscr{E}(obj)$
and $\mathscr{E}(pos)$ are expected to be negatively correlated.
Concretely, for entries at their corresponding locations, if one is
large, the other is desirable to be small. We define the correlation
coefficient in Eq.~\ref{eq:decoupling} between $\mathscr{E}(obj)$
and $\mathscr{E}(pos)$ to be the decouple-ness of $n$ units on the
layer $l$, more negative it is, the better units are pose/identity
decoupled.

%\vspace{-1em}

\begin{equation}
\gamma=corrcoef\,(\mathscr{E}(obj),\mathscr{E}(pose))\label{eq:decoupling}
\end{equation}

\begin{figure}[!htbh]
\begin{centering}
\includegraphics[width=0.48\textwidth]{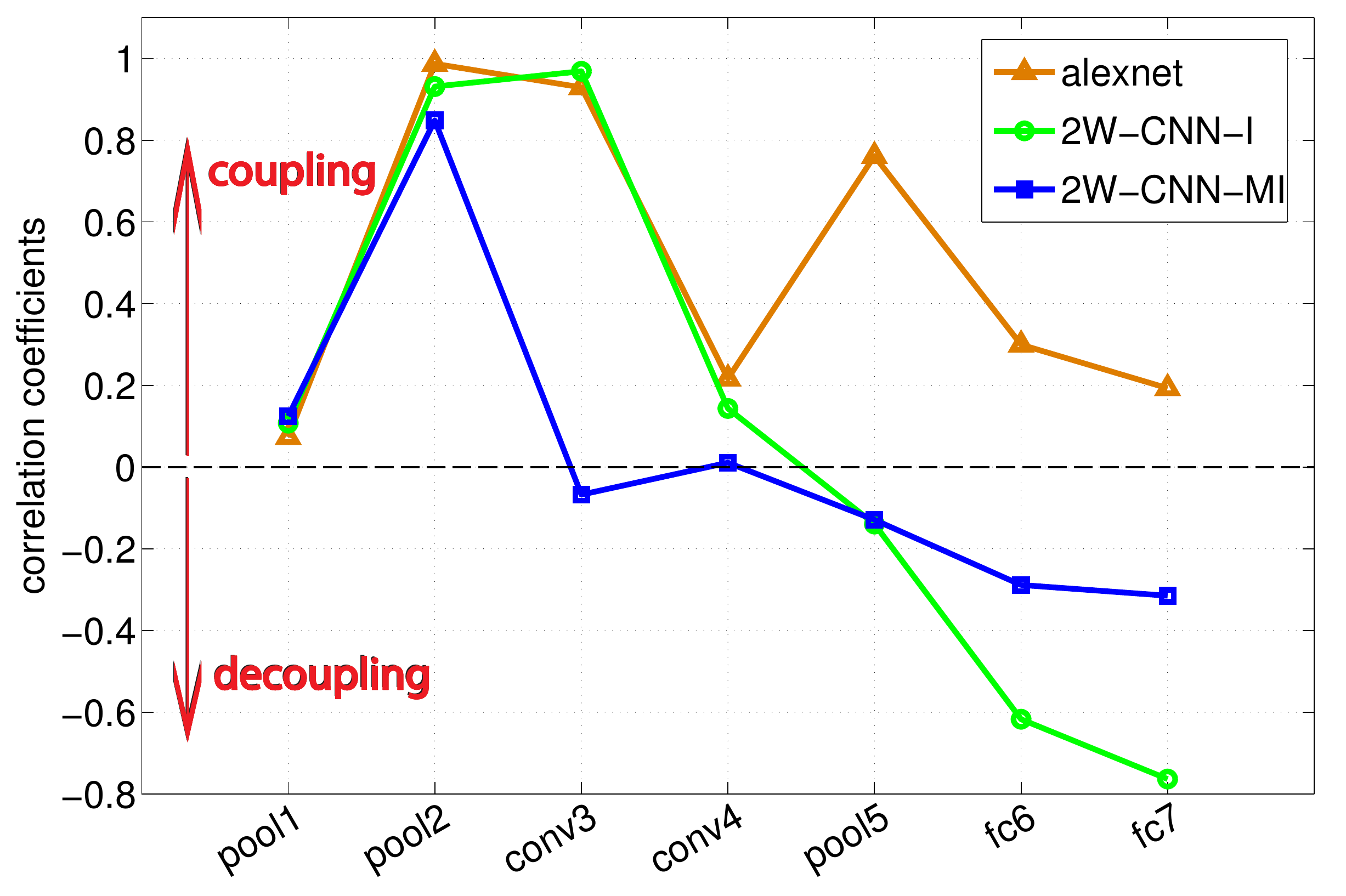}
\par\end{centering}

\caption{\label{fig:decoupling}Decoupling of what and where. This figure shows
the pose/identity decouple-ness of units from the same layer. 2W-CNN
makes pose and identity better decoupled than AlexNet, which indicates
neurons at intermediate layers of 2W-CNN implicitly segregate what
and where information.}
\end{figure}

We compare the decouple-ness of units from our 2W-CNN architecture
against those from AlexNet. We take all units from the same layer,
including pool1, pool2, conv3, conv4, pool5, fc6 and fc7, compute
their decouple-ness and plot them in Fig. \ref{fig:decoupling}. It
reveals: (1) units in 2W-CNN from different layers have been better
what/where decoupled, some are learned to capture pose, while others
are learned to capture identity; (2) units in the earlier layers (e.g.,
pool2, conv3, conv4) are better decoupled in 2W-CNN-MI than in 2W-CNN-I,
which is expected since pose errors are directly back-propagated into
these earlier layers in 2W-CNN-MI. This indicates as well pose and
identity information are implicitly expressed by different units,
although we have no explicit what/where neurons in 2W-CNN.

%\begin{figure*}[htbp!]
%\begin{minipage}[c]{0.70\textwidth}%
%\begin{center}
%\includegraphics[width=0.8\textwidth]{fig-decoupling} 
%\par\end{center}%
%\end{minipage}%\hspace{0.1cm}
%\begin{minipage}[c]{0.30\linewidth}%
%\parbox[t]{1\textwidth}{%
%
%\caption{\label{fig:decoupling}Decoupling of what and where. This figure shows
%the pose/identity decouple-ness of units from the same layer. 2W-CNN
%makes pose and identity better decoupled than AlexNet, which indicates
%neurons at intermediate layers of 2W-CNN implicitly segregate what
%and where information.}
%%
%} %
%\end{minipage}
%\end{figure*}

%There is evidence \cite{malisiewicz2011ensemble} that
%by partitioning the training space according to pose first, training an ensemble of pose-dependent classifiers achieves better classification accuracy
%than training a single classifier on the entire training space. Our
%2W-CNN is different in the way that we do not explicitly
%partition the pose space, but use pose information to guide the training
%process, and in the end, the trained CNNs automatically reconcile
%between pose and identity.

\subsection{Feature visualizations}

%\begin{figure*}[htbp!]
%\begin{minipage}[c]{0.70\textwidth}%
%\begin{center}
%\includegraphics[width=0.8\textwidth]{fig-tsne} 
%\par\end{center}%
%\end{minipage}%\hspace{0.1cm}
%\begin{minipage}[c]{0.30\linewidth}%
%\parbox[t]{1\textwidth}{%
%
%\caption{\label{fig:t-SNE-visualization}t-SNE visualization of fc7 features.
%The learned deep features at fc7 are better delineated by 2W-CNN than
%AlexNet.}
%%
%} %
%\end{minipage}
%\end{figure*}

%\begin{figure}[!htbh]
%\begin{centering}
%\includegraphics[width=0.41\textwidth]{fig-diag-dominant} 
%\par\end{centering}
%
%\centering{}\caption{\label{fig:diag-dominant}Between-class dot-product. We use all test
%images as inputs, compute their fc7 representations, and calculate
%between-class dot-product, which is defined as the median of between-instances
%dot-products of representations. A higher dot-product indicates a
%higher collinearity of their fc7 representations, while a smaller
%dot-product indicates more orthogonality. We further measure the diagonality
%of the dot-product matrix by the ratio of mean within-class dot-product
%over mean between-class dot-product, and the diagonality of (a) AlexNet
%and (b) 2W-CNN-MI is 0.22 and 0.37, showing 2W-CNN-MI achieves more
%desirable representations (more orthogonality). The diagonality measurements
%of their fc6 representations are 0.22 and 0.27 respectively.}
%\end{figure}

\begin{figure}[!htbh]
\includegraphics[width=0.48\textwidth]{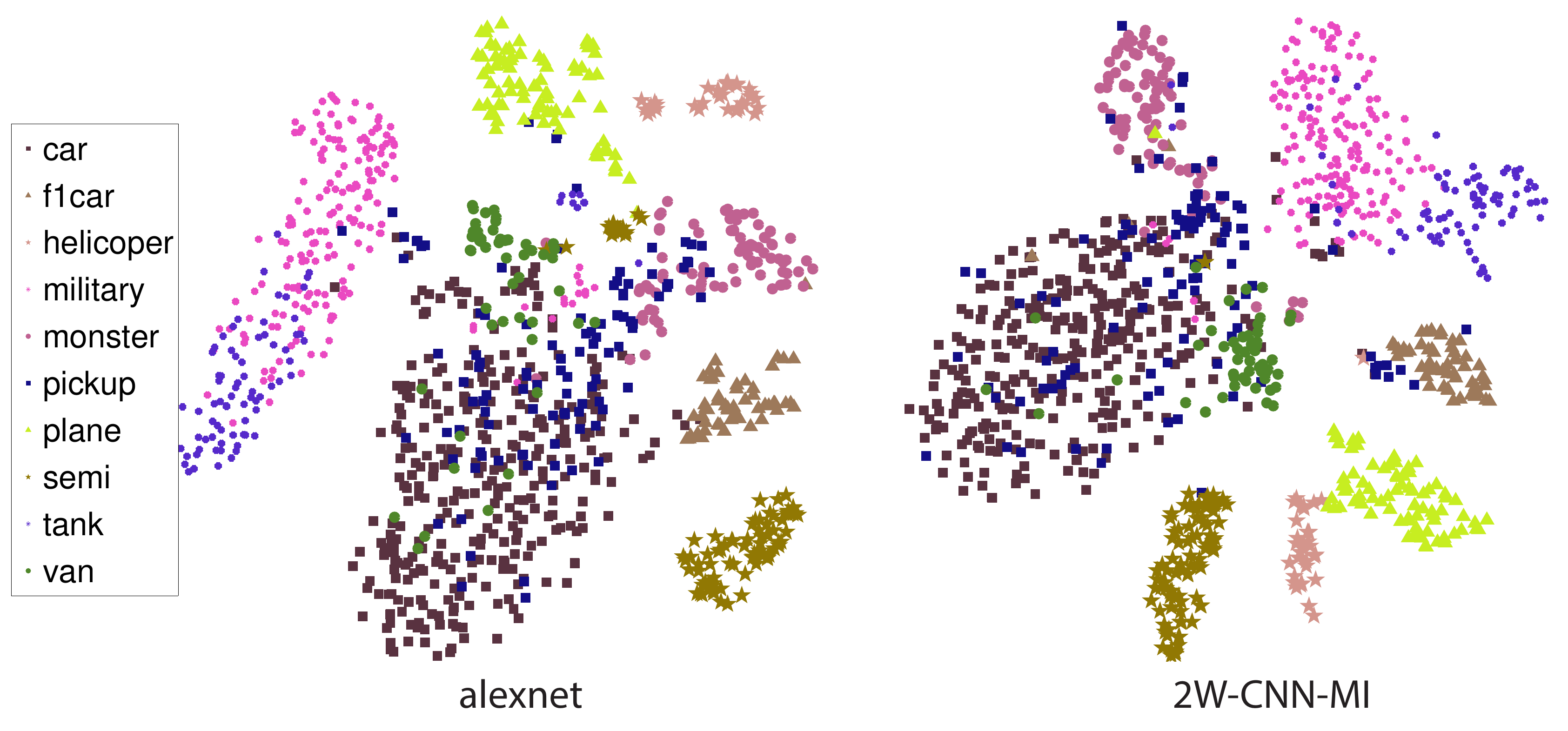}

\caption{\label{fig:t-SNE-visualization}t-SNE visualization of fc7 features.
The learned deep features at fc7 are better delineated by 2W-CNN than
AlexNet.}
\end{figure}

We extract 1024-dimensional features from fc7 of 2W-CNN and AlexNet
as image representations, and use t-SNE \cite{van2008visualizing}
to compute their 2D embeddings and plot results in Fig.~\ref{fig:t-SNE-visualization}.
Seen qualitatively, object categories are better separated by 2W-CNN
representations: For example, ``military car'' (magenta pentagram) and
``tank'' (purple pentagram) representations under 2W-CNN have a clear
boundary, while their AlexNet representation distributions penetrate
into each other. Similarly, ``van'' (green circle) and ``car'' (brown
square) are better delineated by 2W-CNN as well.

\begin{figure*}[!htbh]
\centering{}\includegraphics[width=0.92\textwidth]{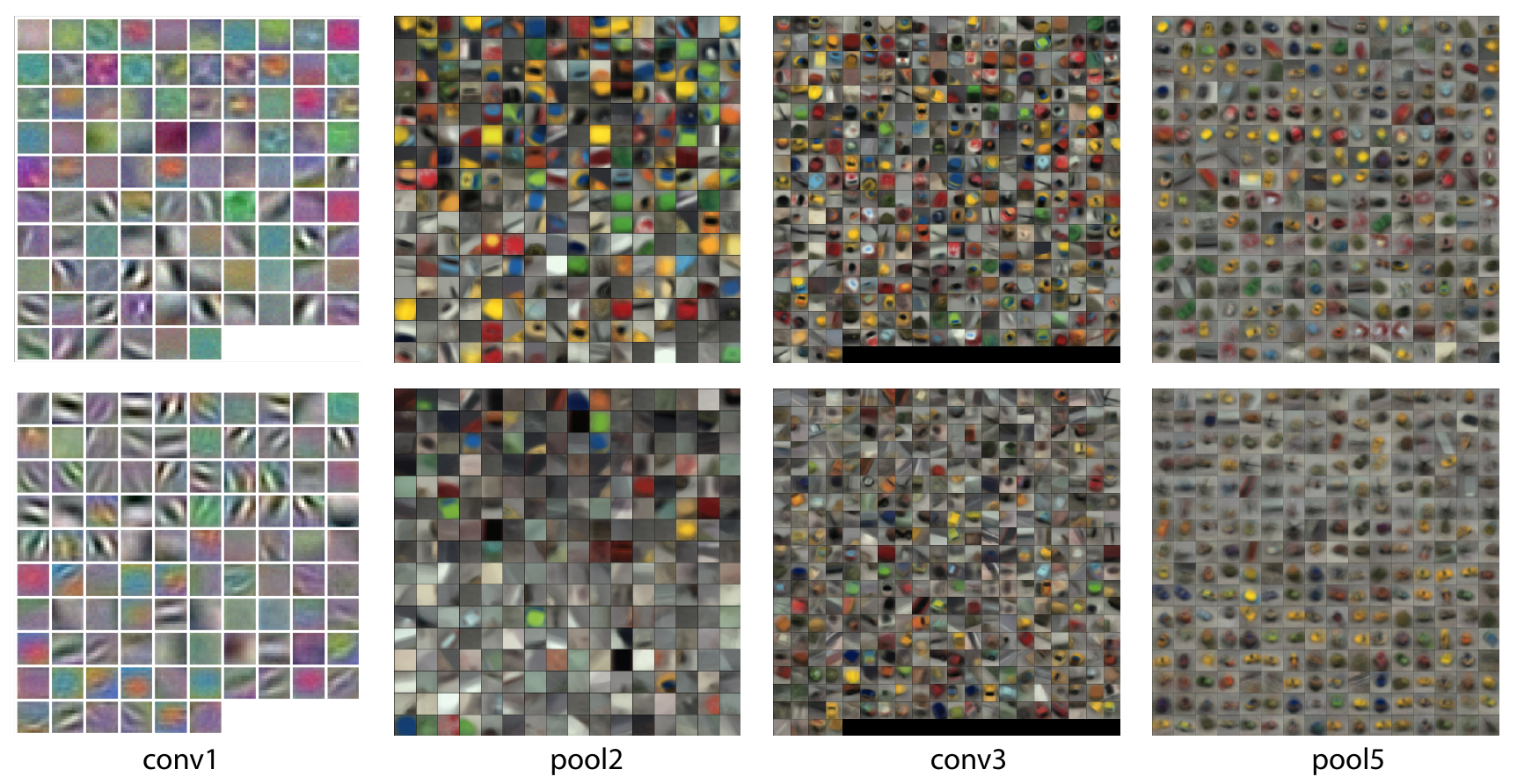}\caption{\label{fig:Visualization-of-receptive}Visualization of receptive
fields of units at different layers. The top (bottom) row shows receptive
fields of AlexNet (2W-CNN). }
\end{figure*}

We further visualize receptive fields of units at different layers
of AlexNet and 2W-CNN. The filters of conv1 can be directly visualized,
while to visualize RFs of units on other layers, we adopt methods
used in \cite{zhou2014object}: we use all test images as input, compute
their activation responses of each unit on each layer, and average
the top 100 images with the strongest activations as a receptive field
visualization of each unit. Fig.~\ref{fig:Visualization-of-receptive}
shows the receptive fields of units on conv1, poo2, conv3 and pool5
of two architectures, AlexNet on top and 2W-CNN-MI on bottom. It suggests
qualitatively that: (1) 2W-CNN has more distinctive and fewer dead
filters on conv1; (2) AlexNet learns many color filters, which can
be seen especially from conv1, pool2 and conv3. While color benefits
object recognition in some cases, configural and structural information
is more desirable in most cases. 2W-CNN learns more structural filters.

\subsection{Extension to ImageNet object recognition}

ImageNet has millions of labeled images, and thus pre-training a ConvNet on another dataset has been shown to yield
insignificant effects \cite{hinton2012deep,lecun2015deep}. To show that the pretrained 2W-CNN-MI and AlexNet on iLab-20M
learns useful features for generic object recognition, we fine-tune the learned weights on ImageNet when we can only
access a small amount of labeled images. We fine-tune AlexNet using 5, 10, 20, 40 images per class from the ILSVRC-2010
challenge. AlexNet is trained and evaluated on ImageNet under three cases: (1) from scratch (use random Gaussian
initialization), (2) from pretrained AlexNet on iLab-20M, (3) from pretrained 2W-CNN-MI on iLab-20M. When we pretrain
2W-CNN-MI on the iLab-20M dataset, we set the units on fc6 and fc7 back to 4096. AlexNet used in pretraining and
finetuning follows exactly as the one in \cite{krizhevsky2012imagenet}. We report top-5 object recognition accuracies in
Table \ref{tab:Top-5-object-recognition-ImageNet}.

%\vspace{-1em}

\begin{table}[tbph]
%\centering{}%
\resizebox{0.48\textwidth}{!}{
\newcommand{\tabincell}[2]{\begin{tabular}{@{}#1@{}}#2\end{tabular}}
\begin{tabular}{|c|c|c|c|c|}
\hline 
\# of images/class  & 5  & 10  & 20  & 40 \tabularnewline
\hline 
\hline
\tabincell{c} {AlexNet \\ (scratch)}  & 1.47  & 4.15  & 16.45  & 25.89 \tabularnewline
\hline 
\tabincell{c} {AlexNet \\ (AlexNet-iLab20M)}  & 7.74  & 12.54  & 19.42  & 28.75 \tabularnewline
\hline 
\tabincell{c} {AlexNet \\ (2W-CNN-MI-iLab20M)}  & \textbf{9.27}  & \textbf{14.85}  & \textbf{23.14}  & \textbf{31.60} \tabularnewline
\hline 
\end{tabular}
}
\caption{\label{tab:Top-5-object-recognition-ImageNet}Top-5 object recognition
accuracies ($\%$) on the test set of ILSVRC-2010, with 150 images
per class and a total of 150K test images. First, fine-tuning AlexNet
from the pretrained features on the iLab-20M dataset clearly outperforms training AlexNet
from scratch, which shows features learned on the iLab-20M dataset
generalizes to ImageNet as well. Second, fine-tuning from the pretrained
2W-CNN-MI (2W-CNN-MI-iLab20M) performs even better than from the pretrained
AlexNet (AlexNet-iLab20M), which shows our 2W-CNN-MI architecture learns
even more effective features for object recognition than AlexNet.}

\end{table}

%\vspace{-2em}

\begin{figure*}
\begin{centering}
\includegraphics [width=0.92\textwidth] {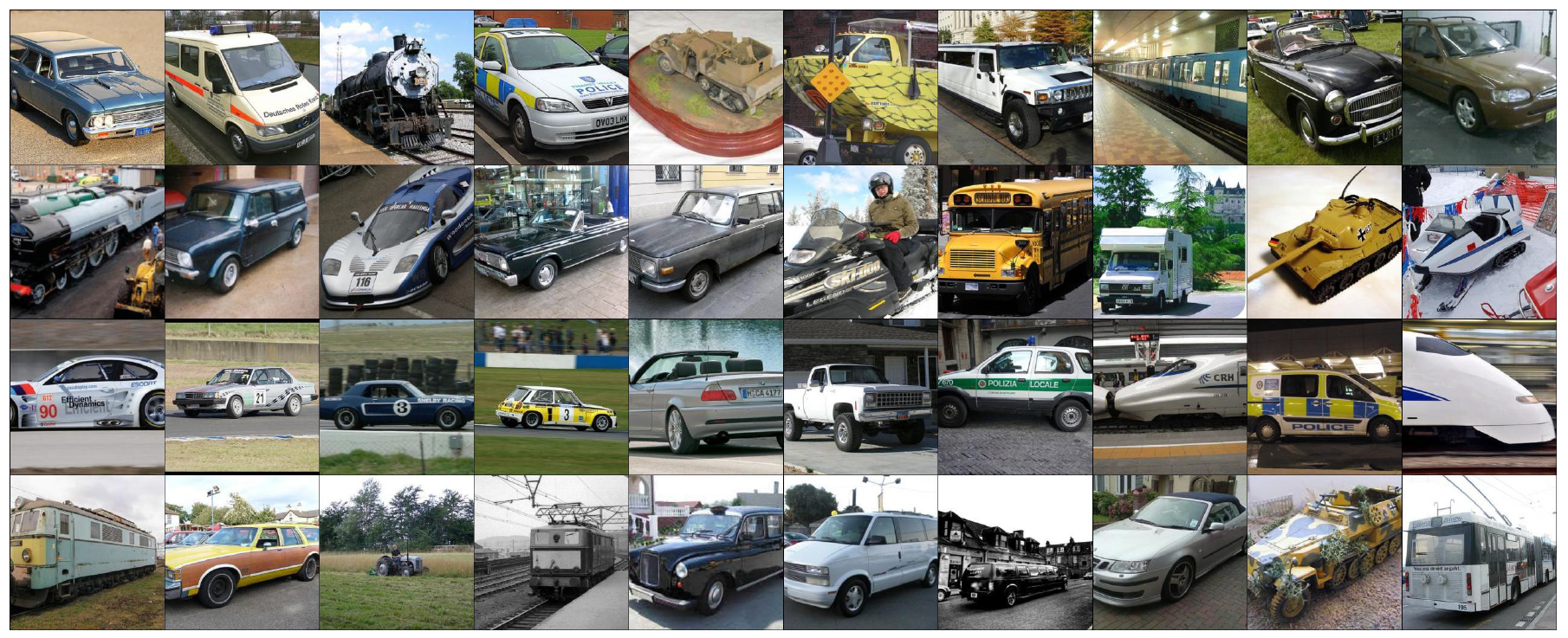}
\par\end{centering}

\centering{}\caption{\label{fig:Pose-estimation-of}Pose estimation of ImageNet images
using trained 2W-CNN-MI on the iLab-20M dataset. Given a test image,
2W-CNN-Mi trained on the iLab-20M dataset could predict one discrete
pose (out of 88). In the figure, each row shows the top 10 vehicle
images which have the same predicted pose label. Qualitatively, images
on the same row do have similar view points, showing 2W-CNN-MI generalizes
well to natural images, even though it is trained on our turntable
dataset.}
\end{figure*}

Quantitative results: we have two key observations (1) when a limited
number of labeled images is available, fine-tuning AlexNet from the
pretrained features on the iLab-20M dataset outperforms training AlexNet
from scratch, e.g., the relative improvement is as large as $\sim530\%$
when we have only 5 samples per class, when more labeled images are
available, the improvement decreases, but we still achieve $\sim22\%$
improvements when 40 labeled images per class are used. This clearly
shows features learned on the iLab-20M dataset generalize well to the
natural image dataset ImageNet. (2) fine-tuning from the pretrained
2W-CNN-MI on iLab-20M performs even better than from the pretrained
AlexNet on iLab-20M, and this shows that 2W-CNN-MI learns even better
features for general object recognition than AlexNet. These empirical
results show that training object categories jointly with pose information
makes the learned features more effective.

Qualitative results: the trained 2W-CNN-MI on iLab-20M could predict
object pose as well; here, we directly use the trained 2W-CNN-MI to
predict pose for each test image from ILSVRC-2010. Each test
image is assigned a pose label (one out of 88 discrete poses, in our
case) with some probability. For each discrete pose, we choose 10
vehicles, whose prediction probabilities at that pose are among the
top 10, and visualize them in Fig. \ref{fig:Pose-estimation-of}.
Each row in Fig. \ref{fig:Pose-estimation-of} shows top 10 vehicles
whose predicted pose label are the same, and as observed, they do
have very similar camera viewpoints. This qualitative result shows
pose features learned by 2W-CNN-MI generalize to ImageNet as well.

%\vspace{-3em}

\section{Conclusion}

Although in experiments we built 2W-CNN on AlexNet, we could use any feed-forward architecture as a base. Our results show that better training can be achieved when explicit
absolute pose information is available. We further show that the pretrained AlexNet and 2W-CNN features on iLab-20M
generalizes to the natural image dataset ImageNet, and moreover, the pretrained 2W-CNN features are shown to be
advantageous to the pretrained AlexNet features in real dataset as well. We believe that this is an important finding
when designing new datasets to assist training of object recognition algorithms, in complement with the existing large
test datasets and challenges.

\noindent \textbf{Acknowledgement}: This work was supported by the National
Science Foundation (grant numbers CCF-1317433
and CNS-1545089), the Army Research Office (W911NF-
12-1-0433), and the Office of Naval Research (N00014-13-
1-0563). The authors affirm that the views expressed herein
are solely their own, and do not represent the views of the
United States government or any agency thereof.

{\small \bibliographystyle{ieee}
\bibliography{objectRecognition-singleStream}
 } 
\end{document}